\documentclass[sigconf, language=english]{acmart}

\usepackage{balance}
\AtBeginDocument{%
  \providecommand\BibTeX{{%
    \normalfont B\kern-0.5em{\scshape i\kern-0.25em b}\kern-0.8em\TeX}}}

\setcopyright{acmlicensed}
\copyrightyear{2018}
\acmYear{2018}

\acmConference[HRI ’24 Causal-HRI Workshop]{HRI ’24: Workshop on Causal Learning for Human-Robot Interaction (Causal -HRI), ACM/IEEE International Conference on Human-Robot Interaction (HRI2024)}{March, 2024}{Boulder, CO.}
\acmBooktitle{HRI ’24: Workshop on Causal Learning for Human-Robot Interaction (Causal -HRI), HRI2024}

\begin{document}

\title{ROS-Causal: A ROS-based Causal Analysis Framework for\\Human-Robot Interaction Applications}

\author{Luca Castri}
\affiliation{%
  \institution{University of Lincoln, UK}\country{}
}
\email{lcastri@lincoln.ac.uk}

\author{Gloria Beraldo}
\affiliation{%
  \institution{University of Padua, Italy}
  \institution{National Research Council of Italy}\country{}
}
\email{gloria.beraldo@cnr.it}

\author{Sariah Mghames}
\affiliation{%
  \institution{University of Lincoln, UK}\country{}
  }
\email{smghames@lincoln.ac.uk}

\author{Marc Hanheide}
\affiliation{%
  \institution{University of Lincoln, UK}\country{}
  }
\email{mhanheide@lincoln.ac.uk}

\author{Nicola Bellotto}
\affiliation{%
  \institution{University of Padua, Italy}
  \institution{University of Lincoln, UK}\country{}
}
\email{nbellotto@dei.unipd.it}


\begin{abstract}
Deploying robots in human-shared spaces requires understanding interactions among nearby agents and objects. Modelling cause-and-effect relations through causal inference aids in predicting human behaviours and anticipating robot interventions. However, a critical challenge arises as existing causal discovery methods currently lack an implementation inside the ROS ecosystem, the standard de facto in robotics, hindering effective utilisation in robotics. To address this gap, this paper introduces ROS-Causal, a ROS-based framework for onboard data collection and causal discovery in human-robot spatial interactions. An ad-hoc simulator, integrated with ROS, illustrates the approach's effectiveness, showcasing the robot onboard generation of causal models during data collection. ROS-Causal is available on GitHub: \url{https://github.com/lcastri/roscausal.git}.
\end{abstract}


\begin{teaserfigure}
  \includegraphics[trim={0cm 0cm 4.5cm 9.3cm},clip,width=\textwidth]{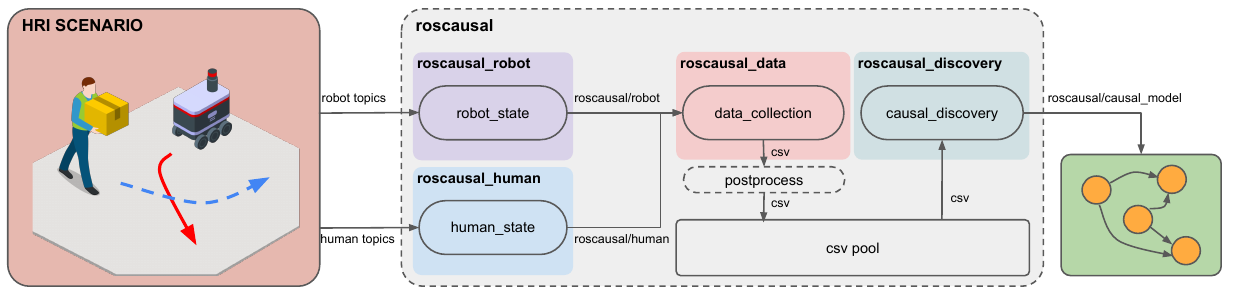}
  \caption{
  ROS-Causal pipeline: \emph{(i)} data extraction from human-robot interaction scenarios; \emph{(ii)} collection and post-processing of data to derive a high-level representation of the scenario. \emph{(iii)} causal discovery conducted on the extracted data, with the resulting causal model published on a dedicated rostopic.}
  \Description{ROS-Causal: a causal analysis framework for human-robot interactions applications.}
  \label{fig:intro}
\end{teaserfigure}

\maketitle

\section{Introduction}
The growing use of robots in various sectors, such as industrial, agriculture, and healthcare, represents an advancement in the development of these sectors. However, to guarantee the robots' inclusion alongside humans in an eco-friendly manner, new approaches for ensuring effective human-robot interactions (HRIs) have been investigated. 
A robot operating near humans must perform tasks with awareness on the unforeseen responses to its actions. 
Uncovering the relationship between robots' actions and their effects on humans can represent a key factor in enhancing HRIs, as it enables the robot to reason about its action. The study of the cause-and-effect relationship is precisely the focus of causal inference~\cite{pearl2009causality}. 
%

Causal inference
appears in the literature of different fields, including robotics~\cite{brawer_causal_2021,cao_reasoning_2021,castri2022causal,castri2023enhancing,Katz2018,Angelov2019,Lee2022,cannizzaro2023towards,cannizzaro2023towardsdrones,cannizzaro2023car}. However, many causal discovery methods used in these applications require two steps: data collection and subsequent offline causal analysis.
Notably, these methods lack the capability to run directly on the robot. This limitation pose challenges for exploiting the reconstructed causal models in real-time. For example, consider a robot interacting with a person in a warehouse, as depicted in Figure~\ref{fig:intro}. In the current scenario, due to the aforementioned limitation, the robot must accumulate a significant amount of data and then conduct offline causal analysis. Subsequently, the reconstructed causal model has to be reintegrated into the robot for utilisation. One reason of such limitation could be the missing of a framework that facilitates the holistic integration between the two communities and that operates directly inside Robot Operating System~(ROS)\footnote{\url{https://www.ros.org/}}, the standard de facto in robotics.
%
The solutions proposed in this paper aim to streamline this process by enabling the robot to conduct onboard causal discovery on data batches while concurrently collecting data for future causal analysis. Moreover, given the integration of our framework within ROS, the acquired causal model could be directly employed by the robot.

Hence, this paper proposes {\em ROS-Causal} to overcome current limitations in causal analysis for real-world robotics applications, and {\em ROS-Causal\_HRISim}, a Gazebo-based robotic simulator useful for designing HRI scenarios for causal analysis. In summary, our contributions are as follows:
\begin{itemize}
    \item the first ROS-based causal analysis framework designed for onboard data collection and causal discovery on robots;
    \item an ad-hoc simulator for human-robot interactions to facilitate the design of HRI scenarios and to collect observational and interventional data for causal analysis;
    \item an experimental evaluation of the proposed approach within the simulated environment to demonstrate its feasibility.
\end{itemize}

The paper is structured as follows: a complete overview of causal discovery methods and their applications in robotics are presented in Section~\ref{sec:related}; Section~\ref{sec:appr} explains the details of our approach; Section~\ref{sec:exp} shows the details of our ad-hoc HRI simulator and the application of our approach; finally, we conclude this paper in Section~\ref{sec:conclusion} discussing achievements and future applications.

\section{Related Work}\label{sec:related}
\textbf{Causal discovery:} Various causal discovery methods have been developed to infer causal relationships from observational data, broadly classified into three categories~\cite{glymour_review_2019}: {\em (i) constraint-based methods}, like Peter \& Clark~(PC) and Fast Causal Inference~(FCI)~\cite{spirtes2000causation}; {\em (ii) score-based methods}, suce as Greedy Equivalence Search~(GES) and NOTEARS~\cite{zheng2018dags}; and {\em (iii) noise-based methods}, like Linear Non-Gaussian Acyclic Models~(LiNGAM)~\cite{shimizu2006linear}.
%
However, many of these algorithms work only with static data and are not applicable to time-series of sensor data in many robotics applications, for which time-dependent causal discovery methods are instead necessary.
To this end, several causal discovery algorithm for time-series data have been developed~\cite{assaad2022survey}. 
Within the area of Granger causality, there is Temporal Causal Discovery Framework~(TCDF)~\cite{nauta2019causal}.
In \cite{pamfil2020dynotears} the time-series version of NOTEARS, i.e. DYNOTEARS, is presented.
Among the noise-based methods, there are Time Series Models with Independent Noise~(TiMINo)~\cite{peters2013causal} and Vector Autoregressive LiNGAM~(VARLiNGAM)~\cite{hyvarinen2010estimation}. In the constraint-based category, variations of the FCI and PC algorithms, namely Time-series FCI~(tsFCI)~\cite{entner2010causal} and PC Momentary Conditional Independence~(PCMCI)~\cite{runge_causal_2018}, were tailored to handle time-series data. 
PCMCI, with wide applications in climate, healthcare, and robotics~\cite{runge_detecting_2019,saetia_constructing_2021,castri2022causal}, has recently seen extensions such as PCMCI\textsuperscript{+}~\cite{runge2020discovering} for discovering simultaneous dependencies and Filtered-PCMCI~(F-PCMCI)~\cite{castri2023enhancing}, which incorporates a transfer entropy-based feature-selection module to enhance causal discovery by focusing on relevant variables.\\
\textbf{Causal robotics:}
The synergy between causality and robotics is a mutually beneficial relationship. Causality utilises robots' physical nature for interventions, while robots use causal models for a deep understanding of their environment. For this reason, causal inference has gained attention in various applications in robotics, such as building Structural Causal Models~(SCM) to understand how humanoid robots interact with tools~\cite{brawer_causal_2021}. PCMCI and F-PCMCI are applied to derive the causal model of an underwater robot reaching a target position~\citep{cao_reasoning_2021} and predict human spatial interactions in social robotics~\citep{castri2022causal,castri2023enhancing}. Causality-based approaches have also been explored in robot imitation learning, manipulation, drone applications, and planning~\cite{Katz2018,Angelov2019,Lee2022,cannizzaro2023towards,cannizzaro2023towardsdrones,cannizzaro2023car}.
However, existing causal discovery methods often involve offline analysis after data collection and are not available in ROS, posing challenges for their use and experimental reproducibility in robotics. Our goal was to create a modular ROS-based causal analysis framework that facilitates concurrent data collection and causal analysis processes.

In robotics, designing an effective HRI is a challenging task. Moreover, the synergy with causality introduces even more variables to be decided, such as features to be considered and how to post-process them. For this reason,
having a simulator that helps in choosing the variables to consider in real-life setting is crucial.
In~\cite{ahmed2020causalworld}, a robotic simulator named CausalWorld, designed for causal structure learning, was introduced. It focuses on a robotic manipulation environment with a TriFinger robot, floor, and stage, and allows for the inclusion of objects with various shapes, such as cubes. While supporting diverse manipulation tasks, it is limited to that domain and lacks the human factor. In contrast, our ROS-Causal\_HRISim is tailored for HRI scenarios and provides the opportunity to collect observational data and perform various types of interventions with both the robot and the human.

\section{ROS-based Causal Analysis Framework}\label{sec:appr}
Our approach, named ROS-Causal, extracts and collects data from an HRI scenario, such as agents' trajectories, and then performs causal analysis on the collected data in a batched manner. A modular ROS Python library implementation of ROS-Causal has been developed and made publicly available\footnote{\url{https://github.com/lcastri/roscausal.git}}. The modular design allows for the expansion of the library with new causal discovery methods. In the following, we provide a detailed explanation of the three main blocks that compose the ROS-Causal pipeline, as depicted in Figure~\ref{fig:intro}. Information regarding subscribers and publishers for each ROS node is summarised in Table~\ref{tab:roscausal}.

\subsection{Data Merging}
The purpose of this block is to merge robot and human data from various topics into custom ROS messages
in the ROS-Causal framework. The nodes \texttt{roscausal\_robot} and \texttt{roscausal\_human} extract the position, orientation, velocities and target positions of the robot and the human, respectively. These data are retrieved from ROS topics/params relative to the robotic platform and need to be configured within the framework. Then, the two nodes merge the acquired data into the ROS messages \texttt{RobotState} and \texttt{HumanState} published on the predefined topics \texttt{/roscausal/robot} and \texttt{/roscausal/human}. The latter are utilised in the data collection block explained in the following section.

\subsection{Data Collection and Post-processing}
The data collection and post-processing block takes input from the previous block's topics to create a data batch for the causal discovery node. More in detail, the \texttt{roscausal\_data} node subscribes to the topics \texttt{/roscausal/robot} and \texttt{/roscausal/human} and begins collecting data in a CSV file. Once the desired time-series length, configurable as a ROS parameter, is reached, the node provides the option to post-process the data, allowing for the creation of a high-level representation of the scenario. For instance, from the low-level data, such as agents' trajectories, a post-processing script can be specified to generate distances and angles between the agents. Once the post-processing is complete, the CSV file is saved into a designated folder (e.g. "csv\_pool" as shown in Figure~\ref{fig:intro}).

\subsection{Causal Discovery}
The \texttt{roscausal\_discovery} ROS node performs causal discovery analysis on the collected data. Specifically, the ROS node continuously checks for the presence of a CSV file in the designated folder. Upon locating a file, it initiates the causal analysis on that specific data batch. If multiple CSV files are present, they are processed and deleted sequentially. The node prioritises the oldest file for the analysis, ensuring that data is analysed in the order it was collected.
It is important to note that the \texttt{roscausal\_data} and \texttt{roscausal\_discovery} ROS nodes operate asynchronously, allowing the simultaneous execution of causal analysis on one dataset while continuing the collection of another. 

The ROS node incorporates two causal discovery methods: the PCMCI~\cite{runge_causal_2018} and its extension, F-PCMCI~\cite{castri2023enhancing}. F-PCMCI has been specifically designed to speed up the causal analysis for real-world scenarios and has shown improved performance compared to PCMCI. For both algorithms, the following parameters, handled as ROS parameters, needs to be set:
\begin{itemize}
    \item significance threshold (typically $\alpha = 0.05$);
    \item minimum and maximum time lag;
    \item conditional independence test;
\end{itemize}
Once the causal analysis is complete, \texttt{roscausal\_discovery} ROS node deletes the just examined CSV dataset in order to maintain robot's memory free and decomposes the causal model into three \texttt{n.lags~$\times$~n.vars~$\times$~n.vars} matrices. Here, \texttt{n.lags} represents the number of time lags to the current time where causal dependencies are tested, defined as the difference between the maximum and minimum time lag, and \texttt{n.vars} represents the number of variables. Each matrix contains distinct information about the built causal model for each time lag. In particular:
\begin{itemize}
    \item \texttt{causal\_structure}: a binary matrix describing the causal model skeleton. Each element is set to $1$ if and only if a causal link between two variables is present;
    \item \texttt{val\_matrix}: specifies the strength of each causal link in the causal model. Notably, this matrix contains non-zero values exclusively for elements where \texttt{causal\_structure} is $1$;
    \item \texttt{pval\_matrix}: indicates the confidence level (i.e., p-value) for each causal link in the causal model. Similar to \texttt{val\_matrix}, this matrix contains positive values only for elements where \texttt{causal\_structure} is set to $1$.
\end{itemize}
The three matrices are incorporated into the \texttt{CausalModel} ROS message and published on the \texttt{/roscausal/causal\_model} topic, making the causal model data accessible to other components of the robotic system.

\begin{table}[t]
  \caption{ROS-Causal subscribers and publisher.}
  \label{tab:roscausal}
  \begin{tabular}{ccc}
    \toprule
    \large\textbf{\texttt{roscausal\_robot}}\\
    \toprule
    \textbf{subscribed topics} & \textbf{description} &\textbf{msg type}\\
    \midrule
    to be setup & robot pose & to be setup\\
    to be setup & robot velocity & to be setup\\
    to be setup & robot goal & to be setup\\
  \bottomrule
    \textbf{published topics} & \textbf{description} &\textbf{msg type}\\
    \midrule
    \texttt{/roscausal/robot} & full robot state & RobotState\\
      \bottomrule
    \toprule
    \large\textbf{\texttt{roscausal\_human}}\\
    \toprule
    \textbf{subscribed topics} & \textbf{description} &\textbf{msg type}\\
    \midrule
    to be setup & human pose & to be setup\\
    to be setup & human velocity & to be setup\\
    to be setup & human goal & to be setup\\
  \bottomrule
    \textbf{published topics} & \textbf{description} &\textbf{msg type}\\
    \midrule
    \texttt{/roscausal/human} & full human state & HumanState\\
      \bottomrule
    \toprule
    \large\textbf{\texttt{roscausal\_data}}\\
    \toprule
    \textbf{subscribed topics} & \textbf{description} &\textbf{msg type}\\
    \midrule
    \texttt{/roscausal/robot} & full robot state & RobotState\\
    \texttt{/roscausal/human} & full human state & HumanState\\
  \bottomrule
    \toprule
    \large\textbf{\texttt{roscausal\_discovery}}\\
    \toprule
    \textbf{published topics} & \textbf{description} &\textbf{msg type}\\
    \midrule
    \texttt{/roscausal/causal\_model} & \begin{tabular}[t]{@{}l@{}}causal model \\ description\end{tabular} & CausalModel\\
  \bottomrule
\end{tabular}
\end{table}
\section{Experiment}\label{sec:exp}
\subsection{Human-Robot Interaction Simulator}
\begin{figure}[t]\centering
\includegraphics[trim={4cm 3.5cm 4cm 4cm}, clip, width=\columnwidth]{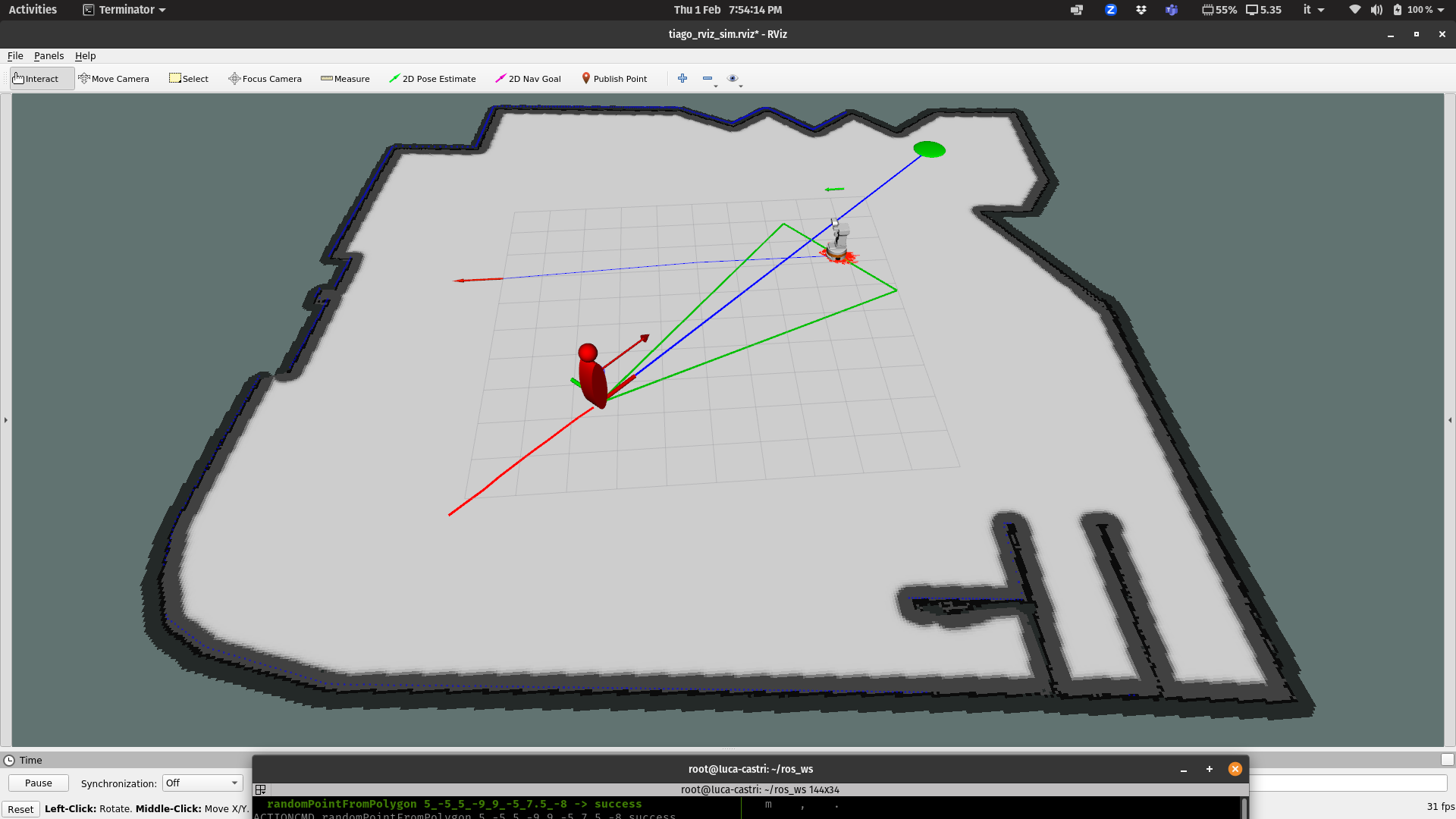}
\caption{HRI scenario involving a TIAGo robot and a teleoperated person, created by ROS-Causal\_HRISim.}
\label{fig:exp_causal_hri}
\end{figure}

To assess the effectiveness of our approach in reconstructing causal models from HRI scenarios, we developed a dedicated Gazebo-based simulator called ROS-Causal\_HRISim. This simulator accurately mimics HRI scenarios involving a TIAGo\footnote{\url{https://pal-robotics.com/robots/tiago/}} robot and multiple pedestrians modelled using the {\em pedsim\_ros}\footnote{\url{https://github.com/srl-freiburg/pedsim_ros}} ROS library. The latter simulates individual and group social activities (e.g., walking) using a social force model. To better emulate human behaviours, we incorporated the option for user teleoperation (via keyboard) of a simulated person, not influenced by social forces.
A Docker image of ROS-Causal\_HRISim, comprising also ROS-Causal, has been created and is publicly available\footnote{\url{https://github.com/lcastri/ROS-Causal_HRISim}}.
An HRI scenario created by ROS-Causal\_HRISim is shown in Figure~\ref{fig:exp_causal_hri}.
\subsection{ROS-Causal Evaluation}
\begin{figure}[b]\centering
\includegraphics[trim={3cm 2.2cm 2.5cm 1.5cm}, clip, width=0.8\columnwidth]{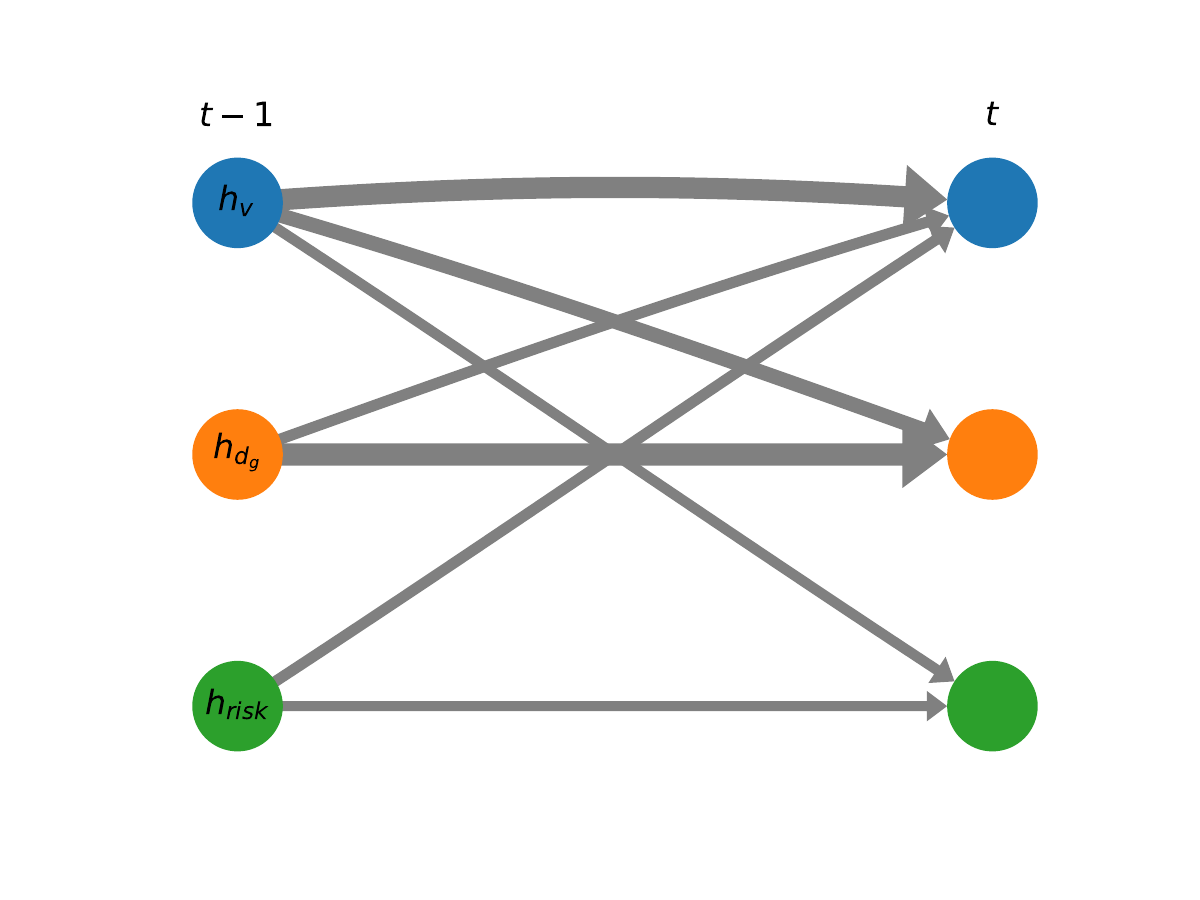}
\caption{Causal model reconstructed by ROS-Causal}
\label{fig:exp}
\end{figure}
To evaluate ROS-Causal, we designed a HRI scenario, depicted in Figure~\ref{fig:exp}, inspired by the scenario analysed in~\cite{castri2022causal}. It involves a TIAGo robot and a teleoperated person, represented by the red manikin. The green dot represents the person's target position, while the blue line visualises the distance between the person and her goal position. Finally, the green cone, built from the person position to the enlarged encumbrance of the TIAGo, which is perceived by the human as a moving obstacle, is the visualisation of the collision risk. The latter is one of the variables defined in~\cite{castri2022causal} and including the following:
\begin{itemize}
    \item $h_v$ - human velocity;
    \item $h_{d_g}$ - distance between the human and his target position;
    \item $h_{risk}$ - risk of collision with the robot.
\end{itemize}
The robot follows a predefined path, while the person, who is teleoperated, has a target position that is randomly chosen within the map and changes once reached by the person. In particular, a new target position is randomly chosen for the person, who then starts moving toward the goal, gradually reducing her velocity as she approaches it. Upon reaching the goal, a new target position is randomly selected, and the process repeats. If, during her path to the goal, the person encounters the robot, she must avoid it by decreasing her velocity and/or adjusting her steering. Therefore, the expected causal links in this scenario are as follows:
\begin{itemize}
    \item $h_v \rightarrow h_{d_g}$, $h_{d_g}$ depends inversely on $h_v$;
    \item $h_{d_g} \rightarrow h_{v} \leftarrow h_{risk}$, $h_{v}$ is a direct function of the $h_{d_g}$, but it is also affected by the collision $risk$;
    \item $h_{v} \rightarrow h_{risk}$ depends on the velocity, as explained in~\cite{castri2022causal}.
\end{itemize}

Regarding the ROS-Causal parameters and settings, we configured a desired time-series length corresponding to a timeframe of $150s$ and recorded the trajectories of the two agents, their linear velocity, and orientation with a sampling time step of $0.3s$. Subsequently, through a dedicated script added to the designated folder in the \texttt{roscausal\_data} ROS node, we post-processed the data to obtain the distance between the human and the goal, as well as the risk of collision. For the causal discovery block, we employed the F-PCMCI causal discovery method with a significance level of $\alpha = 0.05$, a conditional independence test based on Gaussian Process regression and Distance Correlation (GPDC). We also used a 1-step lag time, meaning variables at time $t$ could only be affected by those at time $t-1$. The resulting causal model is depicted in Figure~\ref{fig:exp}, where the thickness of the arrows represents the strength of the causal link. The graph faithfully represents the expected model discussed earlier and is consistent with the results in~\cite{castri2022causal}.

\section{Conclusion}\label{sec:conclusion}
In this work, we introduced {\em ROS-Causal}, a ROS-based causal analysis framework for human-robot interactions applications, and {\em ROS-Causal\_HRISim}, an HRI simulator. ROS-Causal enables onboard data collection and causal discovery, allowing robots to concurrently reconstruct the causal model while collecting data for future causal analysis. Our approach was applied and tested in a simulated HRI scenario designed in ROS-Causal\_HRISim. The results confirm that ROS-Causal can effectively execute data collection and causal discovery analysis onboard, producing accurate causal models.

The current version of ROS-Causal is a fully working yet initial release, with opportunities for future improvements and extensions. Firstly, the ROS nodes, namely \texttt{roscausal\_robot} and \texttt{roscausal\_human}, can be enhanced to accommodate multiple robots and humans. Another promising direction involves the integration of additional causal discovery methods beyond PCMCI and F-PCMCI. This can be easily achieved by introducing new scripts within the dedicated folder for causal discovery methods in the \texttt{roscausal\_discovery} ROS node. There is obviously the opportunity in the future to enhance its compatibility with ROS2. Another possible enhancement is the addition of a new block to the pipeline for leveraging and reasoning on the reconstructed causal models, e.g. \texttt{roscausal\_reasoning}. In any case, implementing the causal analysis as a ROS node and transmitting the causal model via a ROS topic opens up the potential for extensive utilisation of such analyses in the field of robotics. This approach provides the opportunity to leverage causal models for diverse tasks and robots, including but not limited to planning and prediction, in real-time.

\begin{acks}
This work has received funding from the European Union’s Horizon 2020 research
and innovation programme under grant agreement No 101017274 (DARKO).
GB is also supported by PNRR MUR project PE0000013-FAIR.
\end{acks}

\balance
\bibliographystyle{ACM-Reference-Format}
\bibliography{references}


\begin{thebibliography}{26}


\ifx \showCODEN    \undefined \def \showCODEN     #1{\unskip}     \fi
\ifx \showDOI      \undefined \def \showDOI       #1{#1}\fi
\ifx \showISBNx    \undefined \def \showISBNx     #1{\unskip}     \fi
\ifx \showISBNxiii \undefined \def \showISBNxiii  #1{\unskip}     \fi
\ifx \showISSN     \undefined \def \showISSN      #1{\unskip}     \fi
\ifx \showLCCN     \undefined \def \showLCCN      #1{\unskip}     \fi
\ifx \shownote     \undefined \def \shownote      #1{#1}          \fi
\ifx \showarticletitle \undefined \def \showarticletitle #1{#1}   \fi
\ifx \showURL      \undefined \def \showURL       {\relax}        \fi
\providecommand\bibfield[2]{#2}
\providecommand\bibinfo[2]{#2}
\providecommand\natexlab[1]{#1}
\providecommand\showeprint[2][]{arXiv:#2}

\bibitem[Ahmed et~al\mbox{.}(2020)]%
        {ahmed2020causalworld}
\bibfield{author}{\bibinfo{person}{Ossama Ahmed}, \bibinfo{person}{Frederik Tr{\"a}uble}, \bibinfo{person}{Anirudh Goyal}, \bibinfo{person}{Alexander Neitz}, \bibinfo{person}{Manuel Wuthrich}, \bibinfo{person}{Yoshua Bengio}, \bibinfo{person}{Bernhard Sch{\"o}lkopf}, {and} \bibinfo{person}{Stefan Bauer}.} \bibinfo{year}{2020}\natexlab{}.
\newblock \showarticletitle{CausalWorld: A Robotic Manipulation Benchmark for Causal Structure and Transfer Learning}. In \bibinfo{booktitle}{\emph{International Conference on Learning Representations}}.
\newblock


\bibitem[Angelov et~al\mbox{.}(2019)]%
        {Angelov2019}
\bibfield{author}{\bibinfo{person}{Daniel Angelov}, \bibinfo{person}{Yordan Hristov}, {and} \bibinfo{person}{Subramanian Ramamoorthy}.} \bibinfo{year}{2019}\natexlab{}.
\newblock \showarticletitle{{Using causal analysis to learn specifications from task demonstrations}}. In \bibinfo{booktitle}{\emph{Proc. of the Int. Joint Conf. on Autonomous Agents and Multiagent Systems, AAMAS}}.
\newblock


\bibitem[Assaad et~al\mbox{.}(2022)]%
        {assaad2022survey}
\bibfield{author}{\bibinfo{person}{Charles~K Assaad}, \bibinfo{person}{Emilie Devijver}, {and} \bibinfo{person}{Eric Gaussier}.} \bibinfo{year}{2022}\natexlab{}.
\newblock \showarticletitle{Survey and evaluation of causal discovery methods for time series}.
\newblock \bibinfo{journal}{\emph{Journal of Artificial Intelligence Research}}  \bibinfo{volume}{73} (\bibinfo{year}{2022}), \bibinfo{pages}{767--819}.
\newblock


\bibitem[Brawer et~al\mbox{.}(2020)]%
        {brawer_causal_2021}
\bibfield{author}{\bibinfo{person}{Jake Brawer}, \bibinfo{person}{Meiying Qin}, {and} \bibinfo{person}{Brian Scassellati}.} \bibinfo{year}{2020}\natexlab{}.
\newblock \showarticletitle{A causal approach to tool affordance learning}. In \bibinfo{booktitle}{\emph{IEEE/RSJ Int. Conf. on Intell. Robots \& Systems (IROS)}}. \bibinfo{pages}{8394--8399}.
\newblock


\bibitem[Cannizzaro et~al\mbox{.}(2023a)]%
        {cannizzaro2023towardsdrones}
\bibfield{author}{\bibinfo{person}{Ricardo Cannizzaro}, \bibinfo{person}{Rhys Howard}, \bibinfo{person}{Paulina Lewinska}, {and} \bibinfo{person}{Lars Kunze}.} \bibinfo{year}{2023}\natexlab{a}.
\newblock \showarticletitle{Towards Probabilistic Causal Discovery, Inference \& Explanations for Autonomous Drones in Mine Surveying Tasks}. In \bibinfo{booktitle}{\emph{IEEE/RSJ International Conference on Intelligent Robots and Systems (IROS) 2023 Workshop on Causality for Robotics}}. IEEE.
\newblock


\bibitem[Cannizzaro and Kunze(2023)]%
        {cannizzaro2023car}
\bibfield{author}{\bibinfo{person}{Ricardo Cannizzaro} {and} \bibinfo{person}{Lars Kunze}.} \bibinfo{year}{2023}\natexlab{}.
\newblock \showarticletitle{CAR-DESPOT: Causally-Informed Online POMDP Planning for Robots in Confounded Environments}. In \bibinfo{booktitle}{\emph{IEEE/RSJ International Conference on Intelligent Robots and Systems (IROS) 2023}}. IEEE.
\newblock


\bibitem[Cannizzaro et~al\mbox{.}(2023b)]%
        {cannizzaro2023towards}
\bibfield{author}{\bibinfo{person}{Ricardo Cannizzaro}, \bibinfo{person}{Jonathan Routley}, {and} \bibinfo{person}{Lars Kunze}.} \bibinfo{year}{2023}\natexlab{b}.
\newblock \showarticletitle{Towards a Causal Probabilistic Framework for Prediction, Action-Selection \& Explanations for Robot Block-Stacking Tasks}. In \bibinfo{booktitle}{\emph{IEEE/RSJ International Conference on Intelligent Robots and Systems (IROS) 2023 Workshop on Causality for Robotics}}. IEEE.
\newblock


\bibitem[Cao et~al\mbox{.}(2021)]%
        {cao_reasoning_2021}
\bibfield{author}{\bibinfo{person}{Yu Cao}, \bibinfo{person}{Boyang Li}, \bibinfo{person}{Qian Li}, \bibinfo{person}{Adam Stokes}, \bibinfo{person}{David Ingram}, {and} \bibinfo{person}{Aristides Kiprakis}.} \bibinfo{year}{2021}\natexlab{}.
\newblock \showarticletitle{Reasoning Operational Decisions for Robots via Time Series Causal Inference}. In \bibinfo{booktitle}{\emph{2021 IEEE Int. Conf. on Robotics and Automation (ICRA)}}. \bibinfo{pages}{6124--6131}.
\newblock
\urldef\tempurl%
\url{https://doi.org/10.1109/ICRA48506.2021.9561659}
\showDOI{\tempurl}


\bibitem[Castri et~al\mbox{.}(2022)]%
        {castri2022causal}
\bibfield{author}{\bibinfo{person}{Luca Castri}, \bibinfo{person}{Sariah Mghames}, \bibinfo{person}{Marc Hanheide}, {and} \bibinfo{person}{Nicola Bellotto}.} \bibinfo{year}{2022}\natexlab{}.
\newblock \showarticletitle{Causal discovery of dynamic models for predicting human spatial interactions}. In \bibinfo{booktitle}{\emph{International Conference on Social Robotics}}. Springer, \bibinfo{pages}{154--164}.
\newblock


\bibitem[Castri et~al\mbox{.}(2023)]%
        {castri2023enhancing}
\bibfield{author}{\bibinfo{person}{Luca Castri}, \bibinfo{person}{Sariah Mghames}, \bibinfo{person}{Marc Hanheide}, {and} \bibinfo{person}{Nicola Bellotto}.} \bibinfo{year}{2023}\natexlab{}.
\newblock \showarticletitle{Enhancing Causal Discovery from Robot Sensor Data in Dynamic Scenarios}. In \bibinfo{booktitle}{\emph{2nd Conference on Causal Learning and Reasoning}}.
\newblock


\bibitem[Entner and Hoyer(2010)]%
        {entner2010causal}
\bibfield{author}{\bibinfo{person}{Doris Entner} {and} \bibinfo{person}{Patrik~O Hoyer}.} \bibinfo{year}{2010}\natexlab{}.
\newblock \showarticletitle{On causal discovery from time series data using FCI}.
\newblock \bibinfo{journal}{\emph{Probabilistic graphical models}} (\bibinfo{year}{2010}), \bibinfo{pages}{121--128}.
\newblock


\bibitem[Glymour et~al\mbox{.}(2019)]%
        {glymour_review_2019}
\bibfield{author}{\bibinfo{person}{Clark Glymour}, \bibinfo{person}{Kun Zhang}, {and} \bibinfo{person}{Peter Spirtes}.} \bibinfo{year}{2019}\natexlab{}.
\newblock \showarticletitle{Review of causal discovery methods based on graphical models}.
\newblock \bibinfo{journal}{\emph{Frontiers in genetics}}  \bibinfo{volume}{10} (\bibinfo{year}{2019}), \bibinfo{pages}{524}.
\newblock


\bibitem[Hyv{\"a}rinen et~al\mbox{.}(2010)]%
        {hyvarinen2010estimation}
\bibfield{author}{\bibinfo{person}{Aapo Hyv{\"a}rinen}, \bibinfo{person}{Kun Zhang}, \bibinfo{person}{Shohei Shimizu}, {and} \bibinfo{person}{Patrik~O Hoyer}.} \bibinfo{year}{2010}\natexlab{}.
\newblock \showarticletitle{Estimation of a structural vector autoregression model using non-gaussianity.}
\newblock \bibinfo{journal}{\emph{Journal of Machine Learning Research}} \bibinfo{volume}{11}, \bibinfo{number}{5} (\bibinfo{year}{2010}).
\newblock


\bibitem[Katz et~al\mbox{.}(2018)]%
        {Katz2018}
\bibfield{author}{\bibinfo{person}{Garrett Katz}, \bibinfo{person}{Di~Wei Huang}, \bibinfo{person}{Theresa Hauge}, \bibinfo{person}{Rodolphe Gentili}, {and} \bibinfo{person}{James Reggia}.} \bibinfo{year}{2018}\natexlab{}.
\newblock \showarticletitle{{A novel parsimonious cause-effect reasoning algorithm for robot imitation and plan recognition}}.
\newblock \bibinfo{journal}{\emph{IEEE Trans. on Cognitive and Developmental Systems}} (\bibinfo{year}{2018}).
\newblock


\bibitem[Lee et~al\mbox{.}(2021)]%
        {Lee2022}
\bibfield{author}{\bibinfo{person}{Tabitha~E. Lee}, \bibinfo{person}{Jialiang~Alan Zhao}, \bibinfo{person}{Amrita~S. Sawhney}, \bibinfo{person}{Siddharth Girdhar}, {and} \bibinfo{person}{Oliver Kroemer}.} \bibinfo{year}{2021}\natexlab{}.
\newblock \showarticletitle{Causal Reasoning in Simulation for Structure and Transfer Learning of Robot Manipulation Policies}. In \bibinfo{booktitle}{\emph{2021 IEEE Int. Conf. on Robotics and Automation (ICRA)}}. \bibinfo{pages}{4776--4782}.
\newblock
\urldef\tempurl%
\url{https://doi.org/10.1109/ICRA48506.2021.9561439}
\showDOI{\tempurl}


\bibitem[Nauta et~al\mbox{.}(2019)]%
        {nauta2019causal}
\bibfield{author}{\bibinfo{person}{Meike Nauta}, \bibinfo{person}{Doina Bucur}, {and} \bibinfo{person}{Christin Seifert}.} \bibinfo{year}{2019}\natexlab{}.
\newblock \showarticletitle{Causal discovery with attention-based convolutional neural networks}.
\newblock \bibinfo{journal}{\emph{Machine Learning and Knowledge Extraction}} \bibinfo{volume}{1}, \bibinfo{number}{1} (\bibinfo{year}{2019}), \bibinfo{pages}{19}.
\newblock


\bibitem[Pamfil et~al\mbox{.}(2020)]%
        {pamfil2020dynotears}
\bibfield{author}{\bibinfo{person}{Roxana Pamfil}, \bibinfo{person}{Nisara Sriwattanaworachai}, \bibinfo{person}{Shaan Desai}, \bibinfo{person}{Philip Pilgerstorfer}, \bibinfo{person}{Konstantinos Georgatzis}, \bibinfo{person}{Paul Beaumont}, {and} \bibinfo{person}{Bryon Aragam}.} \bibinfo{year}{2020}\natexlab{}.
\newblock \showarticletitle{Dynotears: Structure learning from time-series data}. In \bibinfo{booktitle}{\emph{International Conference on Artificial Intelligence and Statistics}}. PMLR, \bibinfo{pages}{1595--1605}.
\newblock


\bibitem[Pearl(2009)]%
        {pearl2009causality}
\bibfield{author}{\bibinfo{person}{Judea Pearl}.} \bibinfo{year}{2009}\natexlab{}.
\newblock \bibinfo{booktitle}{\emph{Causality}}.
\newblock \bibinfo{publisher}{Cambridge university press}.
\newblock


\bibitem[Peters et~al\mbox{.}(2013)]%
        {peters2013causal}
\bibfield{author}{\bibinfo{person}{Jonas Peters}, \bibinfo{person}{Dominik Janzing}, {and} \bibinfo{person}{Bernhard Sch{\"o}lkopf}.} \bibinfo{year}{2013}\natexlab{}.
\newblock \showarticletitle{Causal inference on time series using restricted structural equation models}.
\newblock \bibinfo{journal}{\emph{Advances in neural information processing systems}}  \bibinfo{volume}{26} (\bibinfo{year}{2013}).
\newblock


\bibitem[Runge(2018)]%
        {runge_causal_2018}
\bibfield{author}{\bibinfo{person}{Jakob Runge}.} \bibinfo{year}{2018}\natexlab{}.
\newblock \showarticletitle{Causal network reconstruction from time series: From theoretical assumptions to practical estimation}.
\newblock \bibinfo{journal}{\emph{Chaos: An Interdisciplinary Journal of Nonlinear Science}} \bibinfo{volume}{28}, \bibinfo{number}{7} (\bibinfo{year}{2018}).
\newblock


\bibitem[Runge(2020)]%
        {runge2020discovering}
\bibfield{author}{\bibinfo{person}{Jakob Runge}.} \bibinfo{year}{2020}\natexlab{}.
\newblock \showarticletitle{Discovering contemporaneous and lagged causal relations in autocorrelated nonlinear time series datasets}. In \bibinfo{booktitle}{\emph{Conference on Uncertainty in Artificial Intelligence}}. PMLR, \bibinfo{pages}{1388--1397}.
\newblock


\bibitem[Runge et~al\mbox{.}(2019)]%
        {runge_detecting_2019}
\bibfield{author}{\bibinfo{person}{Jakob Runge}, \bibinfo{person}{Peer Nowack}, \bibinfo{person}{Marlene Kretschmer}, \bibinfo{person}{Seth Flaxman}, {and} \bibinfo{person}{Dino Sejdinovic}.} \bibinfo{year}{2019}\natexlab{}.
\newblock \showarticletitle{Detecting and quantifying causal associations in large nonlinear time series datasets}.
\newblock \bibinfo{journal}{\emph{Science advances}} \bibinfo{volume}{5}, \bibinfo{number}{11} (\bibinfo{year}{2019}), \bibinfo{pages}{eaau4996}.
\newblock


\bibitem[Saetia et~al\mbox{.}(2021)]%
        {saetia_constructing_2021}
\bibfield{author}{\bibinfo{person}{Supat Saetia}, \bibinfo{person}{Natsue Yoshimura}, {and} \bibinfo{person}{Yasuharu Koike}.} \bibinfo{year}{2021}\natexlab{}.
\newblock \showarticletitle{Constructing brain connectivity model using causal network reconstruction approach}.
\newblock \bibinfo{journal}{\emph{Frontiers in Neuroinformatics}}  \bibinfo{volume}{15} (\bibinfo{year}{2021}), \bibinfo{pages}{619557}.
\newblock


\bibitem[Shimizu et~al\mbox{.}(2006)]%
        {shimizu2006linear}
\bibfield{author}{\bibinfo{person}{Shohei Shimizu}, \bibinfo{person}{Patrik~O Hoyer}, \bibinfo{person}{Aapo Hyv{\"a}rinen}, \bibinfo{person}{Antti Kerminen}, {and} \bibinfo{person}{Michael Jordan}.} \bibinfo{year}{2006}\natexlab{}.
\newblock \showarticletitle{A linear non-Gaussian acyclic model for causal discovery.}
\newblock \bibinfo{journal}{\emph{Journal of Machine Learning Research}} \bibinfo{volume}{7}, \bibinfo{number}{10} (\bibinfo{year}{2006}).
\newblock


\bibitem[Spirtes et~al\mbox{.}(2000)]%
        {spirtes2000causation}
\bibfield{author}{\bibinfo{person}{Peter Spirtes}, \bibinfo{person}{Clark~N Glymour}, {and} \bibinfo{person}{Richard Scheines}.} \bibinfo{year}{2000}\natexlab{}.
\newblock \bibinfo{booktitle}{\emph{Causation, prediction, and search}}.
\newblock \bibinfo{publisher}{MIT press}.
\newblock


\bibitem[Zheng et~al\mbox{.}(2018)]%
        {zheng2018dags}
\bibfield{author}{\bibinfo{person}{Xun Zheng}, \bibinfo{person}{Bryon Aragam}, \bibinfo{person}{Pradeep~K Ravikumar}, {and} \bibinfo{person}{Eric~P Xing}.} \bibinfo{year}{2018}\natexlab{}.
\newblock \showarticletitle{Dags with no tears: Continuous optimization for structure learning}.
\newblock \bibinfo{journal}{\emph{Advances in neural information processing systems}}  \bibinfo{volume}{31} (\bibinfo{year}{2018}).
\newblock


\end{thebibliography}

\end{document}